\documentclass[10pt,twocolumn,letterpaper]{article}

\usepackage{cvpr}              

\definecolor{darkgreen}{rgb}{0.0, 0.5, 0.0}

\newcommand{\cmark}{\ding{52}}%
\newcommand{\xmark}{\ding{56}}%
\newcommand{\tbf}[1]{\textbf{#1}}
\newcommand{\ul}[1]{\underline{#1}}

\definecolor{cvprblue}{rgb}{0.21,0.49,0.74}
\usepackage[pagebackref,breaklinks,colorlinks,allcolors=cvprblue]{hyperref}
\usepackage{xcolor}
\usepackage{array}
\usepackage[table]{xcolor} 
\newcolumntype{g}{>{\columncolor{gray!20}}c}
\usepackage{pifont}
\usepackage{multirow} %
\usepackage{amssymb} 
\def\paperID{} 
\def\confName{CVPR}
\def\confYear{2026}

\title{BEV-Denoise: Learning Intrinsic Noise for Accurate Bird's-Eye-View Semantic Segmentation}

\author{
Dooseop Choi$^{1,2*}$ \quad
Kyounghwan An$^{1}$ \quad
Kyoung-Wook Min$^{1}$ \\
$^{1}$Electronics and Telecommunications Research Institute (ETRI) \\
$^{2}$University of Science and Technology (UST) \\
{\tt\small \{d1024.choi, mobileguru, kwmin92\}@etri.re.kr }
}

\begin{document}
\maketitle

\def\thefootnote{*}\footnotetext{Corresponding author.}
\def\thefootnote{\arabic{footnote}}

\begin{abstract}
In this paper, we present a framework dubbed \textbf{BEV-Denoise} that estimates and removes intrinsic noise from learned Bird's-Eye-View (BEV) features to achieve accurate BEV semantic segmentation. Inspired by the noise estimation capability of Denoising Diffusion Probabilistic Models (DDPM), we design a UNet-based noise estimation module that learns to estimate the noise from the learned BEV features. The estimated noise is then subtracted from the BEV features and fed to BEV map decoders for the final prediction results. To facilitate supervision for the noise estimation module, we follow a sequential learning paradigm called Task Decomposition (TD) where a pre-trained BEV map autoencoder is employed to train a view transformation (VT) encoder. We share three key insights learned from our intensive experiments that are critical for improved performance. We apply our framework to four existing models, encompassing the three major VT paradigms. Experimental results on a large-scale real-world dataset, nuScenes, demonstrate the effectiveness of our framework.
\end{abstract}


\section{Introduction}
\label{sec:intro}
Recognizing driving environments has played a vital role in autonomous driving, as autonomous vehicles (AVs) can only navigate safely with the aid of accurate recognition results. Ranging sensors such as LIDAR or RADAR have long been deployed to perceive the environments since they offer accurate distance information. The fact that the sensors are usually expensive and lack the information abundant in camera sensors has encouraged researchers to pay more attention to learning a unified representation from surround-view camera images. 

Detecting 3D objects from the images is one of the most actively researched tasks \cite{Xiong_cvpr23, Chen_cvpr23, Shu_iccv23, Wang_cvpr23}. Another promising task called Bird's-Eye-View (BEV) semantic segmentation is to predict multiclass occupancy grid maps (OGM), referred to as BEV maps, from surround-view camera images, which are used for subsequent tasks in autonomous driving such as trajectory prediction or motion planning. Many approaches have been proposed for accurate BEV map prediction in the literature \cite{Philion_eccv20, Liu_eccv22, Zhou_cvpr22, Liu_iccv23, Huang_cvpr23, Fang_cvpr23, Yang_cvpr23, Choi_iros24, Peng_wacv23, Li_aaai23, Pan_cvpr23, Liu_icra23, Hu_eccv22}. 

One major challenge in the BEV map prediction has been how to learn a unified representation, referred to as BEV features, from perspective view (PV) images without knowing the true depth, and many research works have been proposed to tackle the challenge. One of the prevalent paradigms is to predict the depth for each pixel position in the image feature maps and then project the features onto BEV grids using the predicted depth as well as camera parameters \cite{Philion_eccv20}. Another is to take advantage of the cross-attention of Transformer \cite{Vaswani_nips17} to learn the mapping from PV features to BEV features from data \cite{Liu_eccv22, Zhou_cvpr22, Li_eccv22, Liu_iccv23}.
\begin{figure}[t]
\centering
\includegraphics[height=6.0cm]{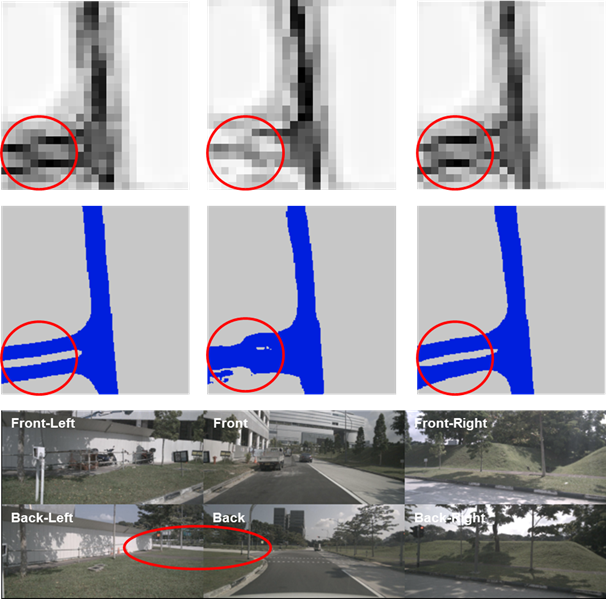}
\caption{Visualization of BEV feature maps (the first row), their corresponding predicted BEV maps for \textit{Drivable Area} (the second row), and input images (the third row). The first, second, and third columns are from the BEV map autoencoder, the VT encoder, and the VT encoder with the proposed denoising network, respectively.}
\label{fig1}
\end{figure}

\begin{figure}[t]
\centering
\includegraphics[height=1.6cm]{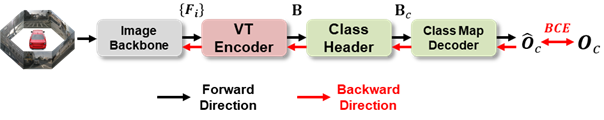}
\caption{General architecture of BEV segmentation models.}
\label{fig1_1}
\end{figure}

BEV features produced by existing BEV segmentation models often suffer from intrinsic noise due to unknown depth, fully or partially occluded objects in the images, and adverse weather conditions as shown in Figure 1. In the figure, the area highlighted by red circles is partially visible to the AV, which hinders the models from producing the clean BEV feature map.

To address this issue, we propose a denoising framework that estimates and removes the intrinsic noise from the learned BEV features. As seen in the figure, the proposed framework effectively estimates and removes the noise, which leads to improved segmentation performance. Our approach is highly inspired by Denoising Diffusion Probabilistic Models (DDPM) \cite{Ho_nips20} where a UNet-based noise estimator is trained to predict the Gaussian noise contaminating the input image. While existing DDPM-based segmentation methods focus on iteratively removing Gaussian noise contaminating the BEV features, hoping that the reverse diffusion process removes the intrinsic noise as well, we are the first to suggest directly estimating and removing the intrinsic noise. However, in the conventional view transformation (VT) encoder and map decoder architecture and its corresponding training method, as shown in Figure \ref{fig1_1}, the ground-truth noise is unavailable during the training. This limitation motivated us to adopt a sequential training paradigm called Task Decomposition (TD) \cite{Zhao_cvpr24}, in which the ground-truth BEV features are obtained by the pre-trained BEV map autoencoder. From extensive experiments, we derived three key insights for improved performance, which will be discussed in detail in later sections. 

In summary, our contributions are the followings:
\begin{itemize}
    \item[$\bullet$] We propose a new framework that estimates and removes intrinsic noise from BEV features to achieve improved segmentation performance.
    \item[$\bullet$] We derive three key lessons from extensive experiments that are critical for improved performance.
    \item[$\bullet$] We apply our framework to four existing baseline models to demonstrate the effectiveness of the proposed approach.
\end{itemize}
\section{Related Works}
\label{sec:related}

\subsection{View Transformation Paradigms}
\label{sec:vt}
One of the major challenges in BEV semantic segmentation is transforming image features in PV space into features in BEV space without access to true depth. Three major paradigms have dominated over the past few years. The first paradigm, often called LSS \cite{Philion_eccv20}, is to predict the depth for each pixel position in the images and use the predicted depth along with camera parameters to lift the image features into the 3D space. The lifted features are then projected onto the BEV grids to obtain the final BEV features. The second paradigm \cite{Liu_eccv22, Zhou_cvpr22} leverages the cross-attention of Transformer \cite{Vaswani_nips17} to intrinsically learn the 2D-to-3D correspondence through BEV queries from data. To mitigate the computational complexity of the second paradigm, the third paradigm \cite{Li_eccv22} proposes replacing the standard cross-attention with deformable cross-attention \cite{Zhu_iclr21}. In this paper, we apply our framework to the four representative methods \cite{Philion_eccv20, Liu_iccv23, Zhou_cvpr22, Li_eccv22} to validate the effectiveness of the proposed approach. 

\subsection{DDPM for BEV Semantic Segmentation}
\cite{Ji_iccv23} first proposed exploiting DDPM for dense prediction tasks, including BEV semantic segmentation. A BEV map decoder is first trained to predict the ground-truth BEV map from both PV image features and a noise-corrupted version of the ground-truth BEV map encoding. During inference, a Gaussian noise map of the same size as the ground-truth BEV map is progressively refined by the decoder via DDIM sampling \cite{Song_iclr21}. \cite{Zou_aaai24} suggested employing DDPM to model the distribution over BEV feature maps learned from a VT encoder. A randomly generated BEV feature map is progressively refined via DDIM sampling with the conditional guidance (e.g., the VT encoder output) and then finally used to update the BEV feature map from the VT encoder. \cite{Le_eccv24} proposed a DDPM-based method to fuse features from different sensors such as LIDAR and camera. Benefiting from the inherent denoising property of the reverse diffusion process, the method is able to refine or even synthesize sensor features, which leads to improved performance. Recently, \cite{Ye_cvpr25} proposed a DDPM-based denoising approach for 3D object detection. Clean BEV feature maps, which are obtained by the standard DDPM denoising process with 3D object labels, are used to regularize noisy BEV feature maps. While the four methods primarily focus on leveraging the generative capabilities of the reverse diffusion process of DDPM, our approach places an emphasis on the noise estimation capability of DDPM. Specifically, we propose deploying a UNet-based network of DDPM to directly estimate intrinsic noise from the learned BEV features in a single forward pass. 

\subsection{Task Decomposition}
\cite{Zhao_cvpr24} first proposed a sequential learning paradigm called Task Decomposition (TD) for frontal camera BEV semantic segmentation. Instead of directly supervising a VT encoder through binary cross entropy between the predicted and ground-truth BEV maps, they proposed training the VT encoder to produce BEV features as close as possible to those from a pre-trained BEV map autoencoder. Finally, the decoder of the autoencoder is fine-tuned using the binary cross entropy to adapt to the learned BEV features. Since the decoder of the autoencoder was initially trained to be robust against noise from a known distribution (e.g., Gaussian) in the BEV space, the combination of the VT encoder and the decoder is expected to be similarly robust to intrinsic noise in the learned BEV features. However, as seen in our experiments, TD does not always guarantee improved performance, suggesting that the intrinsic noise does not necessarily follow the known distribution. In this paper, we follow the TD paradigm to effectively model and estimate the noise from the learned BEV features.

\section{Proposed Approach}

\subsection{Problem Formulation}
Let $I_{i} \in \mathbb{R}^{H_{I} \times W_{I} \times 3}$ denote an image captured by the $i$-th camera mounted on an autonomous vehicle (AV), where $H_{I}$ and $W_{I}$ respectively denote the height and width of the image. Our goal is to predict a BEV map $\mathbf{O} \in \mathbb{R}^{X_{o} \times Y_{o} \times |\mathcal{C}|}$, centered at the AV, from the set of images $\{ I_{i}\}_{i=1}^{N_{c}}$. Here $X_{o}$, $Y_{o}$, and $\mathcal{C}$ denote the width, length, and the set of semantic classes in $\mathbf{O}$, respectively, and $N_{c}$ is the number of the mounted cameras. As shown in Figure \ref{fig1_1}, a general neural network (NN) architecture for the BEV map prediction can be summarized as follows. An image backbone network such as ResNet \cite{He_cvpr16} first processes the input images to produce PV feature maps $ \{F_{i} \in \mathbb{R}^{H_{F} \times W_{F} \times C_{F}} \}_{i}$. Next, a VT encoder transforms them into a BEV feature map $\mathbf{B} \in \mathbb{R}^{X_{B} \times Y_{B} \times C_{B}}$, from which class-specific heads and decoders finally generate the estimate $\hat{\mathbf{O}}$. In general, $X_{B} = \lfloor  X_{o} /2^{s} \rfloor$ and $Y_{B} = \lfloor Y_{o} /2^{s} \rfloor$ hold with a positive integer $s$. 

\subsection{Noise Estimation and Elimination} 
We follow the TD paradigm to effectively model intrinsic noise in the learned BEV feature maps. Let $\mathbf{B}^{ae}$ and $\mathbf{B}$ denote BEV feature maps from a pre-trained BEV map autoencoder and a VT encoder trained under the autoencoder's supervision, respectively. Since $\mathbf{B}^{ae}$ is a high-dimensional representation of $\mathbf{O}$ and the VT encoder was trained to produce $\mathbf{B}$ as close as possible to $\mathbf{B}^{ae}$, we can simply model the noise contamination process as
\begin{equation}
\mathbf{B} = \mathbf{B}^{ae} + \mathbf{N},
\label{eqn1}
\end{equation}
where $\mathbf{N}$ denotes the noise. Our goal is to estimate $\mathbf{N}$ and remove it from $\mathbf{B}$. Highly inspired by the noise estimation capacity of DDPM, we introduce a UNet-based noise estimation network $\mathcal{N}_{\phi}$ that estimates $\mathbf{N}$ as follows:
\begin{equation}
\hat{\mathbf{N}} = \mathcal{N}_{\phi}(\mathbf{B}),
\label{eqn2}
\end{equation}
where $\hat{\mathbf{N}}$ and $\phi$ respectively denote the estimated noise and a set of trainable parameters. The ground-truth for $\hat{\mathbf{N}}$ is directly obtained from Eqn.~\ref{eqn1}, which can be used for supervising $\mathcal{N}_{\phi}$. Let $\mathcal{O}_{NR}$ denote a noise removal operation. Given $\hat{\mathbf{N}}$, a reasonable choice for $\mathcal{O}_{NR}$ is 
\begin{equation}
 \hat{\mathbf{B}}=\mathcal{O}_{NR}(\mathbf{B}, \hat{\mathbf{N}})=\mathbf{B} - \hat{\mathbf{N}}.
\label{eqn3}
\end{equation}
The noise eliminated BEV feature map $\hat{\mathbf{B}}$ is then fed to the decoder of the autoencoder for the final prediction. One can consider other operations for $\mathcal{O}_{NR}$ such as the cross-attention \cite{Zou_aaai24}, which will be discussed in a later section.

\subsection{Key Insights}
In this section, we share key lessons learned from our experiments to improve performance. 

\textbf{1) Joint training of the noise estimation network and the decoder outperforms sequential training.} A natural choice for the loss function to supervise $\mathcal{N}_{\phi}$ is as follows:
\begin{equation}
\mathcal{L}_{noise} = \frac{1}{L}||\mathbf{B}^{ae} - \hat{\mathbf{B}}||^{2}_{2}.
\label{eqn4}
\end{equation}
where $L = X_{B} \times Y_{B} \times C_{B}$. Following the TD paradigm, one may consider training the noise estimation network first and then training the decoder to adjust to the denoised BEV features. However, we found from the experiment in a later section that joint training of the noise estimation network and the decoder yields better performance. The loss in Eqn. \ref{eqn4} encourages $\mathcal{N}_{\phi}$ to focus on reconstructing the missing signal in $\mathbf{B}^{ae}$ from $\mathbf{B}$ more effectively when aided by the BEV map prediction error $BCE(\mathbf{O}, \hat{\mathbf{O}})$ in Eqn. \ref{eqn9}. The main-task supervision from $BCE(\mathbf{O}, \hat{\mathbf{O}})$ provides strong guidance for $\mathcal{N}_{\phi}$, leading to more effective noise estimation aligned with the ultimate goal.

\textbf{2) Simultaneous elimination of both shared and class-specific noise improves the performance.} It is common for BEV segmentation models to predict multi-class information from a shared BEV feature map $\mathbf{B}$. A widely used approach is to introduce class-specific head networks, each extracting a class-specific BEV feature map $\mathbf{B}_{c}$ from $\mathbf{B}$ where $c \in \mathcal{C}$. Following this approach, we modify the architecture of the BEV map autoencoder to have multiple head and decoder networks, each dedicated to a specific class. Even though, in the TD paradigm, the VT encoder is trained to produce $\mathbf{B}$ as close as possible to $\mathbf{B}^{ae}$, the class-specific noise estimation is still performed on class-specific BEV feature maps as follows:
\begin{equation}
\mathbf{B}_{c} = \mathbf{B}_{c}^{ae} + \mathbf{N}_{c},
\label{eqn6}
\end{equation}
\begin{equation}
\hat{\mathbf{N}}_{c} = \mathcal{N}_{\phi}^{cls}(\mathbf{B}_{c}|\mathbf{B}),
\label{eqn7}
\end{equation}
where $\mathbf{B}_{c}^{ae}$ is the BEV feature map obtained from $\mathbf{B}^{ae}$ by the head network for class $c$. Consequently, the class-specific loss is
\begin{equation}
\mathcal{L}_{noise}^{cls} = \frac{1}{L}\sum_{c \in \mathcal{C}}||\mathbf{B}_{c}^{ae} - \hat{\mathbf{B}}_{c}||^{2}_{2},
\label{eqn8}
\end{equation}
where $\hat{\mathbf{B}}_{c} = \mathcal{O}_{NR}(\mathbf{B}_{c}, \hat{\mathbf{N}}_{c})$. Note here that we let $\mathcal{N}_{\phi}^{cls}$ be shared across all classes to avoid a significant increase in the overall network size and complexity. One can consider introducing multiple noise estimation networks, each corresponding to a specific class, for improved performance at the cost of the increased computational complexity and overall network size. 

\textbf{3) Noise estimation network learns shape patterns in BEV map images.} As shown in Table \ref{tab3}, the proposed denoising approach works well on static classes such as \textit{Drivable Area} and \textit{Divider}, whose shapes typically exhibit consistent patterns on BEV map images. On the other hand, it is not effective for \textit{Vehicle} and \textit{Pedestrian} classes. This is because the locations and headings of these objects on BEV map images are typically random, which makes the noise estimation challenging. As a consequence, we propose applying the proposed approach only to the static classes. Unless stated otherwise in later sections, the proposed noise estimation and elimination processes are not applied to \textit{Vehicle} and \textit{Pedestrian}, resulting in performance identical to that of TD.

\begin{figure}[t]
\centering
\includegraphics[height=5.0cm]{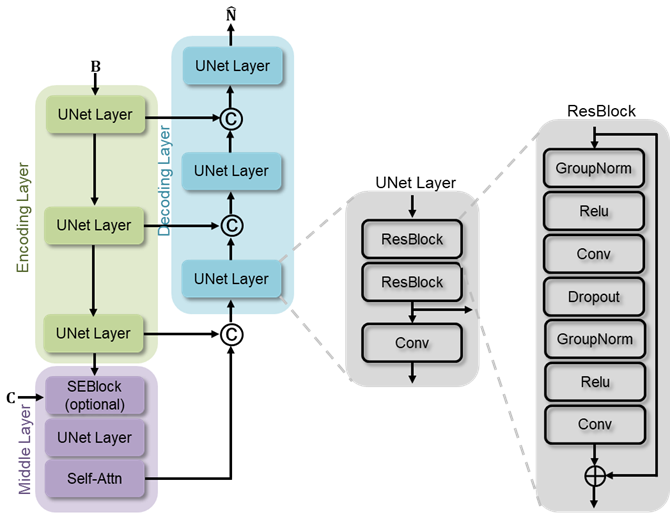}
\caption{The overall architecture of the proposed noise estimation network with $R_{U}=3$.}
\label{fig3}
\end{figure}

\begin{figure}[t]
\centering
\includegraphics[height=6.5cm]{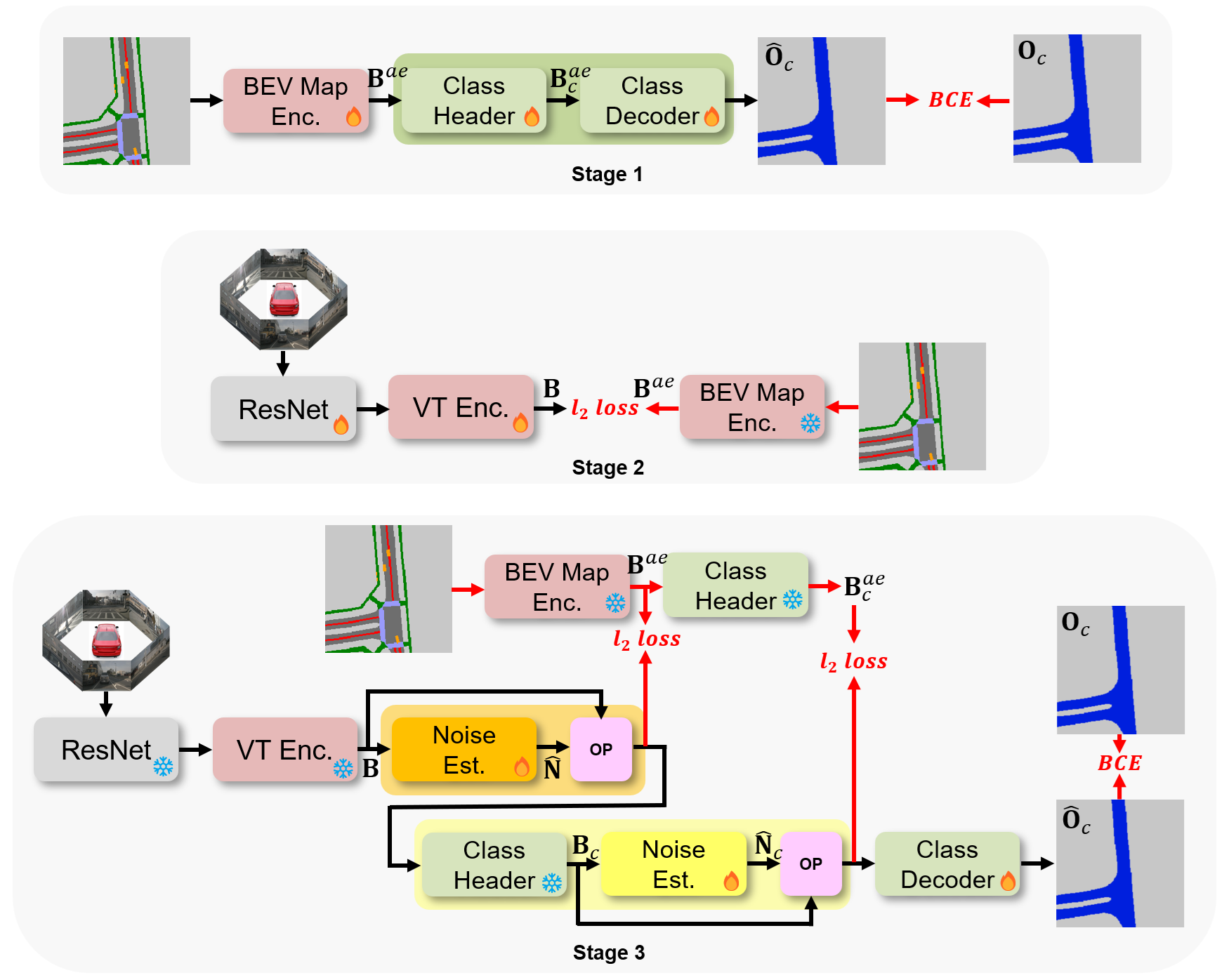}
\caption{The overall architecture of the proposed approach. For the sake of simplicity, we take an example of class \textit{Drivable Area}. For the experiments in later sections, the other classes including \textit{Divider}, \textit{Cross Walk} are simultaneously predicted through the corresponding head and decoder networks. The red arrows are for training purposes only.}
\label{fig2}
\end{figure}

\subsection{Noise Estimation Network Architecture} 
Figure \ref{fig3} shows the overall architecture of the proposed UNet-based noise estimation network, which is inspired by those used in DDPM. The middle layer of the network can optionally have \textbf{SEBlock} \cite{Hu_cvpr18} to condition the input as shown in Eqn. \ref{eqn7}. Finally, whenever \textit{Up-Sampling} or \textit{Down-Sampling} operations are required, we let \textbf{ResBlock} include the corresponding layers. The number of UNet layer repetitions, denoted as $R_{U}$, in the encoding and decoding layers plays an important role in enhancing the segmentation performance. We further analyze its impact on the performance in the ablation study section. Unless stated otherwise, we set $R_{U} = 3$.

\subsection{Overall Training Procedure}
The overall training procedure consists of three stages following the TD paradigm as shown in Figure \ref{fig2}. Note that, for simplicity, we use the example of the \textit{Drivable Area} class in the figure. However, for the experiments in later sections, we predict all seven classes at the same time. We first train a BEV map autoencoder using the ground-truth BEV maps. Next, a VT encoder is trained under the supervision of the autoencoder. Finally, the noise estimation networks (one for shared and the other for class-specific BEV feature maps) are trained using the pre-trained autoencoder and VT encoder. At the same time, the class-specific decoders of the autoencoder are fine-tuned to adapt to the denoised BEV feature maps. The final loss to be optimized at the third stage is then
\begin{equation}
\mathcal{L} = BCE(\mathbf{O},\hat{\mathbf{O}}) + \alpha \cdot \mathcal{L}_{noise} +\beta \cdot \mathcal{L}_{noise}^{cls},
\label{eqn9}
\end{equation}
where $BCE$ denotes the binary cross entropy loss. The predefined constants $\alpha$ and $\beta$ are empirically set to 1. The proposed training procedure is different from the TD paradigm in that the noise estimation networks are simultaneously trained with the decoders at the third stage. 

\begin{table*}[t]
\scriptsize
\begin{center}
\begin{tabular}{|c|c| c c c c c c c|c|}
\hline
Model & $\#$ param. (M) & \textit{Driv.} & \textit{Div.} & \textit{Ped. X} & \textit{Walk.} & \textit{CarPark.} & \textit{Veh.} & \textit{Ped.} & Average\\
\hline
CVT                       & 18.9  & 74.9       & 32.9       & 38.6       & 45.3       & 41.3       & 28.1       & 3.3        & 37.7 \\
CVT+DiffBEV               & 72.3  & 75.1       & 33.0       & 38.6       & 45.9       & \tbf{43.8} & 28.1       & 3.4        & $38.2_{0.5\uparrow}$ \\
CVT+TD                    & 11.5  & 72.7       & 30.5       & 34.8       & 44.5       & 36.6       & \ul{30.8}       & \tbf{9.4}   & $36.9_{0.8\downarrow}$ \\
\rowcolor{gray!10}
CVT+Ours ($R_{U}=1$)      & 47.0  & \ul{76.1}  & \ul{34.6}  & \ul{42.8}  & \ul{49.6}  & 41.8       & \tbf{30.9} & \tbf{9.4}  & $\ul{40.8}_{3.1\uparrow}$ \\
\rowcolor{gray!20}
CVT+Ours ($R_{U}=3$)      & 101.3 & \tbf{77.2} & \tbf{35.2} & \tbf{44.7} & \tbf{50.8} & \ul{42.7}  & \tbf{30.9} & \tbf{9.4}  & $\tbf{41.5}_{3.8\uparrow}$ \\
\hline
PETR                      & 98.5   & 70.7       & 30.6       & \tbf{35.8} & 41.1       & 36.2       & 18.0       & 1.1       & 33.3 \\
PETR+DiffBEV              & 151.8  & 67.8       & 28.7       & 30.0       & 38.0       & 35.1       & 15.5       & 1.3       & $30.9_{2.4\downarrow}$ \\
PETR+TD                   & 91.1   & 71.2       & 30.2       & 32.7       & 41.6       & 36.2       & \tbf{21.6} & \tbf{2.9}  & $33.7_{0.4\uparrow}$ \\
\rowcolor{gray!10}
PETR+Ours ($R_{U}=1$)     & 126.6  & \ul{72.0}  & \ul{30.9}  & 34.0       & \ul{42.5}  & \ul{37.0}  & \ul{21.5}  & \tbf{2.9} & $\ul{34.4}_{1.1\uparrow}$ \\
\rowcolor{gray!20}
PETR+Ours ($R_{U}=3$)     & 180.9  & \tbf{72.1} & \tbf{31.3} & \ul{34.8} & \tbf{42.8}  & \tbf{37.2} & \ul{21.5}  & \tbf{2.9}  & $\tbf{34.6}_{1.3\uparrow}$ \\
\hline
LSS                       & 25.5  & 72.7       & 30.5       & \tbf{37.3} & 43.4       & \tbf{37.5} & 14.4       & 1.3       & 33.8 \\
LSS+DiffBEV               & 79.1  & 71.5       & 30.0       & 35.5       & 42.3       & 36.5       & 13.1       & 1.3       & $32.7_{1.1\downarrow}$ \\
LSS+TD                    & 21.0  & 72.1       & 30.1       & 34.4       & 43.0       & 35.6       & \tbf{15.3} & \tbf{1.5} & $33.1_{0.7\downarrow}$ \\
\rowcolor{gray!10}
LSS+Ours ($R_{U}=1$)      & 57.6  & \ul{73.1}  & \ul{30.9}  & 36.1       & \ul{44.2}  & 37.0       & \ul{15.2}  & \tbf{1.5} & $\ul{34.0}_{0.2\uparrow}$ \\
\rowcolor{gray!20}
LSS+Ours ($R_{U}=3$)      & 111.3 & \tbf{73.4} & \tbf{31.2} & \ul{37.2}  & \tbf{44.8} & \ul{37.3}  & \tbf{15.3} & \tbf{1.5} & $\tbf{34.3}_{0.5\uparrow}$ \\
\hline
BEVFormer                     & 30.60 & \ul{78.3}  & \tbf{41.5} & \tbf{51.1} & \ul{54.4}  & \ul{45.9}  & \tbf{37.5} & 9.6        & \tbf{45.5} \\
BEVFormer+DiffBEV             & 84.20 & \tbf{78.5} & \ul{41.3}  & \ul{49.6}  & \tbf{54.8} & \tbf{47.6} & \ul{36.9}  & 9.6        & $\ul{45.4}_{0.1\downarrow}$\\
BEVFormer+TD                  & 26.10 & 77.2       & 37.6       & 45.1       & 50.5       & 38.7       & 35.3       & 11.9       & $42.3_{3.2\downarrow}$ \\
\rowcolor{gray!10}
BEVFormer+Ours ($R_{U}=1$)    & 62.1  & 77.9       & 39.5       & 47.0       & 51.9       & 43.1       & 35.2       & \ul{11.9}  & $43.8_{1.7\downarrow}$ \\
\rowcolor{gray!20}
BEVFormer+Ours ($R_{U}=3$)    & 116.4 & \ul{78.3}  & 40.2       & 48.4       & 53.0       & 41.2       & 35.2       & \tbf{12.0} & $44.0_{1.5\downarrow}$ \\

\hline
\end{tabular}
\end{center}
\caption{Quantitative comparison on nuScenes. The values are reported in mIoU. The bold and underline indicate the best and second-best performance, respectively. The values in the subscript indicate the performance gain over the baseline.}
\label{tab1}
\end{table*}

\begin{figure*}[t]
\centering
\includegraphics[height=12.0cm, width=17.0cm]{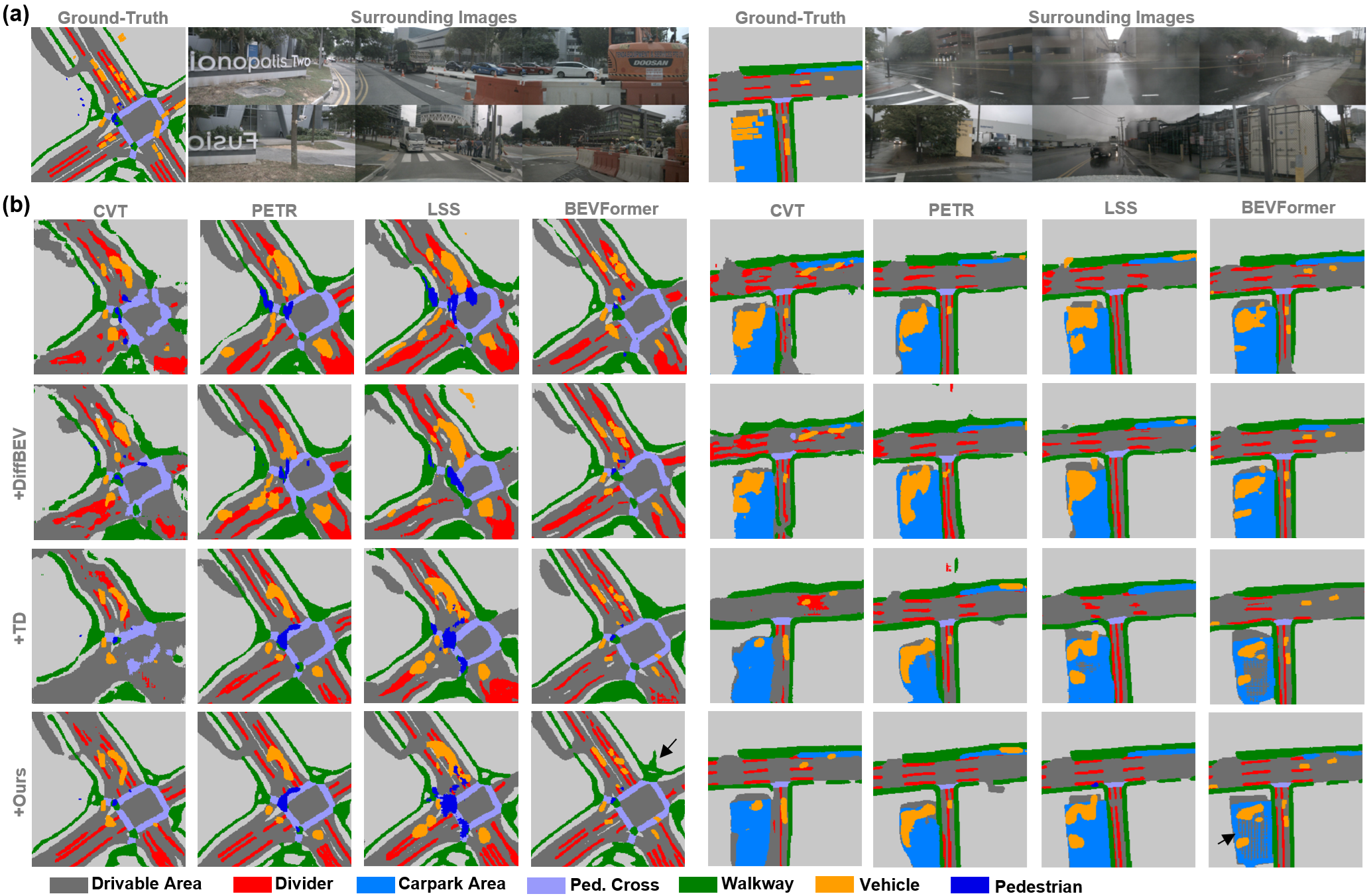}
\caption{Visualization of (a) surround-view camera images and their corresponding ground-truth BEV maps, and (b) BEV map prediction results. In (b), the first row shows the predictions from the four baseline models. The second, third, and fourth rows show the results when DiffBEV, TD, and Ours are applied to the baseline models, respectively.}
\label{fig4}
\end{figure*}

\section{Experiments}
\subsection{Dataset and Evaluation Settings}
To demonstrate the effectiveness of the proposed approach, it is applied to the four representative methods LSS \cite{Philion_eccv20}, CVT \cite{Zhou_cvpr22}, PETR \cite{Liu_iccv23}, and BEVFormer \cite{Li_eccv22}, and is evaluated on a large-scale real-world dataset, nuScenes \cite{Caesar_cvpr20}. Following the standard practice \cite{Zhou_cvpr22, Fang_cvpr23, Choi_iros24}, the ground-truth BEV maps are generated at a resolution of 200 $\times$ 200 pixels by orthographically projecting HD map elements and 3D bounding boxes in a 100m $\times$ 100m area around the ego-vehicle onto the ground plane. The input camera images are scaled and cropped to 224 $\times$ 448 pixels before being fed to an image backbone. The classes considered are \textit{Drivable Area}, \textit{Divider}, \textit{Pedestrian Cross}, \textit{Walkway}, \textit{CarPark Area}, \textit{Vehicle}, and \textit{Pedestrian}. Finally, for the evaluation metric, we use Intersection-over-Union (IoU) score between the ground-truth and its prediction.

\subsection{Network and Training details}
Two image backbones \cite{He_cvpr16, Tan_icml19} are used for the four baselines following the respective original implementations. Following the TD paradigm, we first train two BEV map autoencoders with the initial learning rate of 5$e^{-4}$: one produces $\mathbf{B}^{ae} \in \mathbb{R}^{25 \times 25 \times 256}$ for CVT and PETR, and the other produces $\mathbf{B}^{ae} \in \mathbb{R}^{50 \times 50 \times 256}$ for LSS and BEVFormer. Next, a VT encoder is trained under the supervision of the respective autoencoder with the learning rate reported in the corresponding papers. Finally, the \textit{shared} and \textit{class-specific} noise estimation networks and the decoder of the autoencoder are simultaneously trained via the loss in Eqn. \ref{eqn9} with the initial learning rate of 1$e^{-4}$. In implementing DiffBEV \cite{Zou_aaai24}, because the original implementation from the corresponding authors does not provide the conditional DDPM part for the semantic segmentation task, we implemented it based on the description in the paper. We adapted the noise estimation network architecture ($R_{U}=3$) proposed in our paper for a fair comparison. We incorporated DDIM \cite{Song_iclr21} for fast sampling, and found that the best performance was achieved with four sampling steps. All models are trained with a batch size of 4 for 30 epochs using AdamW \cite{Loshchilov_iclr19} on four NVIDIA RTX 3090 GPUs. More details can be found in the supplementary material.

\subsection{Objective Evaluation}
In Table \ref{tab1}, we compare the four models trained with the proposed approach to those trained with or without the existing approaches, DiffBEV \cite{Zou_aaai24} and TD \cite{Zhao_cvpr24}. We note that, since the proposed approach is based on the TD paradigm and is not applied to \textit{Vehicle} and \textit{Pedestrian}, its performance on these two classes is nearly identical to that of TD. The table shows that neither DiffBEV nor TD consistently improves the performance across all baseline models. Specifically, TD degrades the performance of CVT, LSS, and BEVFormer, with the most noticeable drop observed in BEVFormer. This suggests that TD may be less effective in the surround-view camera setting than in the original front-facing camera setup. DiffBEV, on the other hand, shows noticeable performance improvement only with CVT. We implemented DiffBEV without the depth loss proposed in the corresponding paper, which is one of the key factors contributing to the improved performance, for a fair comparison. With the introduction of the depth loss, better results than those in Table \ref{tab1} may be expected. In contrast, the proposed approach consistently improves all the models except BEVFormer, with the most significant gain observed in CVT. Although the proposed method underperforms on BEVFormer due to its reliance on the TD paradigm, it still outperforms BEVFormer+TD by a large margin.

\subsection{Subjective Evaluation}
We present, in Figure \ref{fig4}, the input images, the respective ground-truth BEV maps, and the prediction results. As shown in the figure, the proposed method effectively reconstructs missing parts of the maps predicted under TD. Specifically, the trained noise estimation networks appear to have prior knowledge of general patterns in static classes such as \textit{Drivable Area} and \textit{Divider}, leading to improved performance. It is worth noting that the missing parts in the prediction by TD are mainly due to the thresholding operation used for visualization. We believe that the weak signals representing these missing parts in the BEV features are recovered by the proposed noise elimination process. On the flip side, this reconstruction ability can sometimes negatively affect predictions, as highlighted by the black arrows in the results of BEVFormer+Ours. As a result, BEVFormer+Ours performs worse than BEVFormer in terms of the mIoU metric, despite generating more plausible results overall. On the other hand, both DiffBEV and TD fail to effectively improve the four baseline models. The trends observed in the figure are also similarly found in other prediction results in the supplementary materials. 

\begin{table}[t]
\begin{center}
\scalebox{0.6}{
\begin{tabular}{|c|c c c|c c c c c|c|}
\hline
Model & Shared & Class & Noise Est. & \textit{Driv.} & \textit{Div.} & \textit{Ped. X} & \textit{Walk.} & \textit{CarPark.} & Avg.\\
\hline
 CVT                              & - & - & -  & 74.9       & 32.9       & 38.6       & 45.3       & 41.3       & 46.6 \\
\rowcolor{gray!20}
\textbf{M1}  & \cmark & \cmark & \cmark  & \tbf{77.2} & \tbf{35.2} & \tbf{44.7} & \tbf{50.8} & 42.7       & \tbf{50.1} \\
\textbf{M2}  & \cmark & \cmark & -       & 76.4       & 33.8       & 41.5       & 48.9       & 42.6       & 48.6 \\
\hline
\textbf{M1a}  & \cmark & \cmark & \cmark  & 76.8       & 35.1       & 44.2       & 50.6       & 41.9       & 49.7 \\
\textbf{M1b}  & \cmark & \xmark & \cmark  & 76.7       & 34.7       & 42.9       & 49.8       & \tbf{43.8} & 49.6 \\
\textbf{M1c}  & \xmark & \cmark & \cmark  & 76.1       & 34.5       & 42.1       & 49.1       & 40.1       & 48.4\\

\hline
\end{tabular}
}
\end{center}
\caption{Ablation study on the effectiveness of our insights. The values are reported in mIoU. When training $\mathbf{M1a}$, the noise estimation network is trained first, followed by fine-tuning the decoder to adapt to the denoised BEV features.}
\label{tab2}
\end{table}

\begin{table}[t]
\begin{center}
\scalebox{0.7}{
\begin{tabular}{|c|c g c|}
\hline
Class               & CVT+TD & \textbf{M1}      & \textbf{M2} \\
\hline
\textit{Drivable}   & 72.7  & \tbf{77.2}   & 76.8 \\
\textit{Divider}    & 30.0  & \tbf{35.2}   & 35.0 \\
\hline
\textit{Vehicle}    & 30.8  & \tbf{30.9}   & 27.3 \\
\textit{Pedestrian} & \tbf{9.4}   & \tbf{9.4}    & 6.1 \\
\hline

\end{tabular}
}
\end{center}
\caption{Effect of the proposed approach on static and dynamic classes. The values are reported in mIoU.}
\label{tab3}
\end{table}

\begin{table}[t]
\begin{center}
\scalebox{0.7}{
\begin{tabular}{|c|c|c|c|c|}
\hline
$R_{U}$ & 1 & 2 & 3 & 4\\
\hline
$\#$ of param. of $\mathcal{N}_{\phi}$ & 17.7& 31.3 & 44.8 & 58.4 \\
mIoU  & 49.0 & 49.9 & 50.1 & 50.6\\

\hline
\end{tabular}
}
\end{center}
\caption{Performance variation of CVT+Ours with respect to the number of U-Net layer repetitions ($R_{U}$). The mIoU values are averaged over static classes, and the number of parameters is reported in millions.}
\label{tab4}
\end{table}

\begin{table}[t]
\begin{center}
\scalebox{0.7}{
\begin{tabular}{|c|c|c|c|}
\hline
& Subtraction & Cross Attn. & Concat.+Conv.\\
\hline
mIoU  & 50.1 & 43.5 & 49.6\\
\hline

\end{tabular}
}
\end{center}
\caption{Ablation study on the choice of the noise removal process ($\mathcal{O}_{NR}$). We use CVT as the baseline model, and the mIoU values are averaged over static classes.}
\label{tab5}
\end{table}

\begin{table}[t]
\begin{center}
\scalebox{0.7}{
\begin{tabular}{|c|c|c|c|c|}
\hline
                   & CVT   & PETR  & LSS   & BEVFormer\\
\hline
-                    & 22.1 & 46.5 & 55.2 & 63.8 \\
w/ DiffBEV ($R_{U}=3$)          & 66.6  & 78.5 & 93.0 & 101.6  \\
w/ Ours ($R_{U}=1$) & 42.5 & 56.9 & 77.1 & 84.3 \\
w/ Ours ($R_{U}=3$) & 52.1 & 66.8 & 87.1 & 96.6 \\

\hline

\end{tabular}
}
\end{center}
\caption{Inference time (ms) per scene. The values are the average across 1000 test scenes.}
\label{tab6}
\end{table}


\subsection{Ablation Study}
We conducted ablation studies to demonstrate the effectiveness of our insights presented in the previous section. The models \textbf{M1} and \textbf{M2} in Table 2 denote CVT+Ours trained with and without the noise estimation losses (Eqn. \ref{eqn4} and \ref{eqn8}), respectively. The results show that the noise estimation networks are effectively trained to remove the intrinsic noise through the proposed loss. It is also found from \textbf{M1} and \textbf{M1a} that \textit{joint training of the noise estimation network and the decoder outperforms sequential training}. On the other hand, we estimate and remove noise using either a \textit{shared} (\textbf{M1b}) or \textit{class-specific} (\textbf{M1c}) noise estimation network alone. The experiments show that \textit{performance improvement is maximized when both shared and class-specific noise estimation networks are applied}, even though each is also effective when used individually. Finally, to verify that \textit{noise estimation network learns shape patterns in BEV map images}, we let the class-specific network estimate and remove the noise not only for static classes (classes with particular shape patterns on BEV map images such as \textit{Drivable Area} and \textit{Divider}) but also for dynamic classes (classes without such patterns such as \textit{Vehicle} and \textit{Pedestrian}). In Table \ref{tab3}, \textbf{M2} and \textbf{M1} denote CVT+Ours applied to all classes and to only static classes, respectively, for the noise estimation and removal process. It shows that the proposed approach is ineffective for dynamic classes. Even though \textit{Vehicle} and \textit{Pedestrian} are depicted as rectangles on the map images, their sizes, locations, and headings can be arbitrary on the images, resulting in inconsistent patterns that are harder to learn. 

Further experiments were conducted using CVT+Ours to show (1) how the performance varies with the size of the noise estimation network, (2) which operation for $\mathcal{O}_{NR}$ yields the best performance, and (3) how the inference speed of the four models varies with the proposed approach. As shown in Table \ref{tab4}, the segmentation performance improves as we increase the size of the noise estimation network, which is natural. Table \ref{tab5} shows that the simple subtraction operation in Eqn. \ref{eqn3} is more effective than the cross-attention proposed in \cite{Zou_aaai24} and the concatenation, which is reasonable given our noise contamination process in Eqn. \ref{eqn1}. Finally, we report the inference time (ms) of each model with or without the proposed approach in Table \ref{tab6}. While the shared noise estimation network processes a single input BEV feature map, the class-specific network must repeat the process $| \mathcal{O} |$ times - once for each class. To eliminate this redundancy, we concatenate the class-specific BEV maps along the batch axis and let the noise estimation network process them at once during the inference. The table shows that increasing the number of layers in the noise estimation network leads to longer inference time. However, it also verifies that the four baseline models can still run under 100ms, which suggests the potential for real-time applications with the proposed method. On the other hand, DiffBEV turned out to be slower than ours because of its iterative sampling during the reverse diffusion process.

\section{Conclusion and Future Works}
In this paper, we proposed a new framework that estimates and removes intrinsic noise from BEV feature maps to improve BEV segmentation models. This framework is highly inspired by the noise estimation capacity of UNet-based networks in DDPM. While existing DDPM-based methods focus on iteratively removing Gaussian noise contaminating BEV feature maps, hoping that the reverse process removes the intrinsic noise as well, we suggested directly estimating the noise and remove it from the BEV feature maps. Since the noise, in general, is unavailable unlike in DDPM, we adopted the TD paradigm to directly access the noise during the training. Through extensive experiments, we learned three lessons to make the proposed approach more effective, which is one of our major contributions. We designed the architecture of the noise estimation network based on those in DDPM, which turned out to be successful. However, further research needs to be done to optimize the network complexity without notably compromising performance.

{
    \small
    \bibliographystyle{ieeenat_fullname}
    \bibliography{main}

@String(AAAI = {AAAI})

@inproceedings{Philion_eccv20,
  author  = "J. Philion and S. Fidler",
  year    = 2020,
  title   = "{Lift, Splat, Shoot: encoding images from arbitrary camera rigs by implicitly unprojecting to 3D}",
  booktitle = "Eur. Conf. Comput. Vis."
}

@inproceedings{Vaswani_nips17,
  author  = "A. Vaswani and N. Shazeer and N. Parmar and J. Uszkoreit and L. Jones and A. N. Gomez and L. Kaiser and I. Polosukhin",
  year    = 2017,
  title   = "{Attention is all you need}",
  booktitle = "Adv. Neural Inform. Process. Syst."
}

@inproceedings{Liu_eccv22,
  author  = "Y. Liu and T. Wang and X. Zhang and J. Sun",
  year    = 2022,
  title   = "{PETR: position embedding transformation for multi-view 3d object detection}",
  booktitle = "Eur. Conf. Comput. Vis."
}

@inproceedings{Zhou_cvpr22,
  author  = "B. Zhou and P. Krahenbuhl",
  year    = 2022,
  title   = "{Cross-view transformers for real-time map-view semantic segmentation}",
  booktitle = "IEEE Conf. Comput. Vis. Pattern Recog."
}

@inproceedings{Li_eccv22,
  author  = "Z. Li and W. Wang and H. Li and E. Xie and C. Sima and T. Lu and Y. Qiao and J. Dai",
  year    = 2022,
  title   = "{BEVFormer: learning bird's-eye-view representation from multi-camera images via spatiotemporal transformers}",
  booktitle = "Eur. Conf. Comput. Vis."
}

@inproceedings{Liu_iccv23,
  author  = "Y. Liu and J. Yan and F. Jia and S. Li and A. Gao and T. Wang and X. Zhang",
  year    = 2023,
  title   = "{PETRv2: A Unified Framework for 3D Perception from Multi-Camera Images}",
  booktitle = "Int. Conf. Comput. Vis."
}

@inproceedings{Zhao_cvpr24,
  author  = "T. Zhao and Y. Chen and Y. Wu and T. Liu and B. Du and P. Xiao and S. Qiu and H. Yang and G. Li and Y. Yang and Y. Lin",
  year    = 2024,
  title   = "{Improving Bird's Eye View Semantic Segmentation by Task Decomposition}",
  booktitle = "IEEE Conf. Comput. Vis. Pattern Recog."
}

@inproceedings{Ho_nips20,
  author  = "J. Ho and A. Jain and P. Abbeel",
  year    = 2020,
  title   = "{Denoising Diffusion Probabilistic Models}",
  booktitle = "Adv. Neural Inform. Process. Syst."
}

@inproceedings{Zhu_iclr21,
author = "X. Zhu and W. Su and L. Lu and B. Li and X. Wang and J. Dai",
year = 2021,
title = "{Deformable detr: deformable transformers for end-to-end object detection}",
booktitle = "Int. Conf. on Learn. Represent."
}

@inproceedings{Zou_aaai24,
author = "J. Zou and Z. Zhu and Y. Ye and X. Wang",
year = 2024,
title = "{DiffBEV: Conditional Diffusion Model for Bird's Eye View Perception}",
booktitle = "the AAAI Conf. Artificial Intell."
}

@inproceedings{Le_eccv24,
  author  = "D.-T. Le and H. Shi and J. Cai and H. Rezatofighi",
  year    = 2024,
  title   = "{DifFUSER: Diffusion Model for Robust Multi-Sensor Fusion in 3D Object Detection and BEV Segmentation}",
  booktitle = "Eur. Conf. Comput. Vis."
}

@inproceedings{Ji_iccv23,
  author  = "Y. Ji and Z. Chen and E. Xie and L. Hong and X. Liu and
Z. Liu and T. Lu and Z. Li and P. Luo",
  year    = 2023,
  title   = "{DDP: Diffusion Model for Dense Visual Prediction}",
  booktitle = "Int. Conf. Comput. Vis."
}

@InProceedings{Song_iclr21,
 author = {J. Song and C. Meng and S. Ermon},
 title = {Denoising Diffusion Implicit Models},
 booktitle = {Int. Conf. on Learn. Represent.},
 year = {2021}
}

@inproceedings{He_cvpr16,
 author = {K. He and X. Zhang and S. Ren and J. Sun},
 title = {Deep residual learning for image recognition},
 booktitle = {IEEE Conf. Comput. Vis. Pattern Recog.},
 year = {2016}
}

@inproceedings{Huang_cvpr23,
 author = {Y. Huang and W. Zheng and Y. Zhang and J. Zhou and J. Lu},
 title = {Tri-Perspective View for Vision-Based 3D Semantic Occupancy Prediction},
 booktitle = {IEEE Conf. Comput. Vis. Pattern Recog.},
 year = {2023}
}

@inproceedings{Fang_cvpr23,
 author = {S. Fang and Z. Wang and Y. Zhong and J. Ge and S. Chen and Y. Wang},
 title = {TBP-Former: learning temporal bird's-eye-view pyramid for joint perception and prediction in vision-centric autonomous driving},
 booktitle = {IEEE Conf. Comput. Vis. Pattern Recog.},
 year = {2023}
}

@inproceedings{Yang_cvpr23,
 author = {C. Yang and Y. Chen and H. Tian and C. Tao and X. Zhu and Z. Zhang and G. Huang and H. Li and Y. Qiao and L. Lu and J. Zhou and J. Dai},
 title = {BEVFormer v2: adapting modern image backbones to bird's-eye-view recognition via perceptive supervision},
 booktitle = {IEEE Conf. Comput. Vis. Pattern Recog.},
 year = {2023}
}

@inproceedings{Pan_cvpr23,
 author = {C. Pan and Y. He and J. Peng and Q. Zhang and W. Sui and Z. Zhang},
 title = {BAEFormer: bi-directional and early interaction transformers for bird's eye view semantic segmentation},
 booktitle = {IEEE Conf. Comput. Vis. Pattern Recog.},
 year = {2023}
}

@inproceedings{Choi_iros24,
author = {D. Choi and J. Kang and T. An and K. An and K. Min},
title = {Progressive Query Refinement Framework for Bird’s-Eye-View Semantic Segmentation from Surrounding Images},
booktitle = {Int. Conf. Intell. Robots Syst.},
year = {2024}
}

@inproceedings{Hu_cvpr18,
 author = {J. Hu and L. Shen and S. Albanie and G. Sun and E. Wu},
 title = {Squeeze-and-Excitation Networks},
 booktitle = {IEEE Conf. Comput. Vis. Pattern Recog.},
 year = {2018}
}

@inproceedings{Caesar_cvpr20,
 author = {H. Caesar and V. Bankiti and A. H. Lang and S. Vora and V. E. Liong and Q. Xu and A. Krishnan and Y. Pan and G. Baldan and O. Beijbom},
 title = {nuScenes: a multimodal dataset for autonomous driving},
 booktitle = {IEEE Conf. Comput. Vis. Pattern Recog.},
 year = {2020}
}

@InProceedings{Tan_icml19,
 author = {M. Tan and Q. V. Le},
 title = {EfficientNet: Rethinking Model Scaling for Convolutional Neural Networks},
 booktitle = {Int. Conf. on Mach. Learn.},
 year = {2019}
}

@InProceedings{Loshchilov_iclr19,
 author = {I. Loshchilov and F. Hutter},
 title = {Decoupled Weight Decay Regularization},
 booktitle = {Int. Conf. on Learn. Represent.},
 year = {2019}
}

@InProceedings{Xiong_cvpr23,
 author = {K. Xiong and S. Gong and X. Ye and X. Tan and J. Wan and E. Ding and J. Wang and X. Bai},
 title = {CAPE: camera view position embedding for multi-view 3d object detection},
 booktitle = {IEEE Conf. Comput. Vis. Pattern Recog.},
 year = {2023}
}

@InProceedings{Chen_cvpr23,
 author = {D. Chen and J. Li and V. Guizilini and R. Ambrus and A. Gaidon},
 title = {Viewpoint equivariance for multi-view 3d object detection},
 booktitle = {IEEE Conf. Comput. Vis. Pattern Recog.},
 year = {2023}
}

@InProceedings{Shu_iccv23,
author = {C. Shu and J. Deng and F. Yu and Y. Liu},
title = {3DPPE: 3D point positional encoding for multi-camera 3D object detection transformers},
booktitle = {Int. Conf. Comput. Vis.},
year = {2023}
}

@InProceedings{Wang_cvpr23,
 author = {Y. Wang and Y. Chen and Z. Zhang},
 title = {FrustumFormer: adaptive instance-aware resampling for multi-view 3D detection},
 booktitle = {IEEE Conf. Comput. Vis. Pattern Recog.},
 year = {2023}
}

@InProceedings {Peng_wacv23,
author = {L. Peng and Z. Chen and Z. Fu and P. Liang and E. Cheng},
booktitle = {2023 IEEE/CVF Winter Conf. on Appli. of Compt. Vis. (WACV)},
title = {BEVSegFormer: Bird's Eye View Semantic Segmentation From Arbitrary Camera Rigs},
year = {2023}
}

@article{Li_aaai23,
  title={BEVDepth: Acquisition of Reliable Depth for Multi-view 3D Object Detection},
  author={Y. Li and Z. Ge and G. Yu and J. Yang and Z. Wang and Y. Shi and J. Sun and Z. Li},
  booktitle={the AAAI Conf. Artificial Intell.},
  year={2023}
}

@InProceedings{Liu_icra23,
 author = {Z. Liu and H. Tang and A. Amini and X. Yang and H. Mao and D. L. Rus and S. Han},
 title = {BEVFusion: multi-task multi-sensor fusion with unified bird's-eye view representation},
 booktitle = {IEEE Int. Conf. Robotics and Automation},
  year = {2023}
}

@InProceedings{Hu_eccv22,
 author = {S. Hu and L. Chen and P. Wu and H. Li and J. Yan and D. Tao},
 title = {ST-P3: end-to-end vision-based autonomous driving via sptio-temporal feature learning},
 booktitle = {Eur. Conf. Comput. Vis.},
 year = {2022}
}

@InProceedings{Ye_cvpr25,
 author = {X. Ye and B. Yaman and S. Cheng and F. Tao and A. Mallik and L. Ren},
 title = {BEVDiffuser: Plug-and-Play Diffusion Model for BEV Denoising with Ground-Truth Guidance},
 booktitle = {arXiv:2502.19694},
 year = {2025}
}
}

\clearpage
\setcounter{page}{1}
\maketitlesupplementary


\begin{figure*}[t]
\centering
\includegraphics[width=15.0cm]{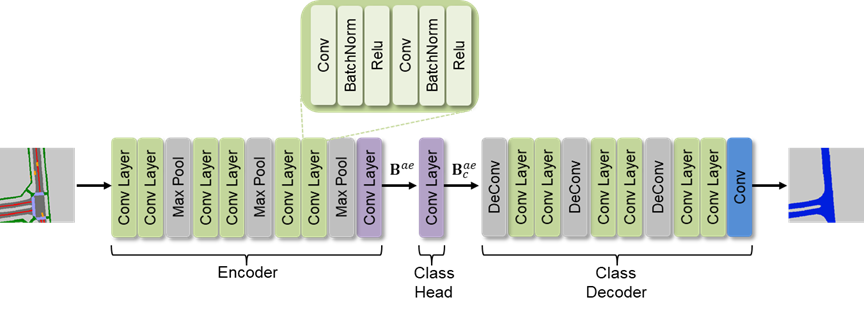}
\caption{The overall architecture of the BEV map autoencoder. Depending on the spatial resolution of $\mathbf{B}^{ae}$, \textbf{MaxPool} and \textbf{DeConv} operations in the encoder and decoder can be dropped.}
\label{sup_fig1}
\end{figure*}

\begin{figure*}[t]
\centering
\includegraphics[width=15.0cm]{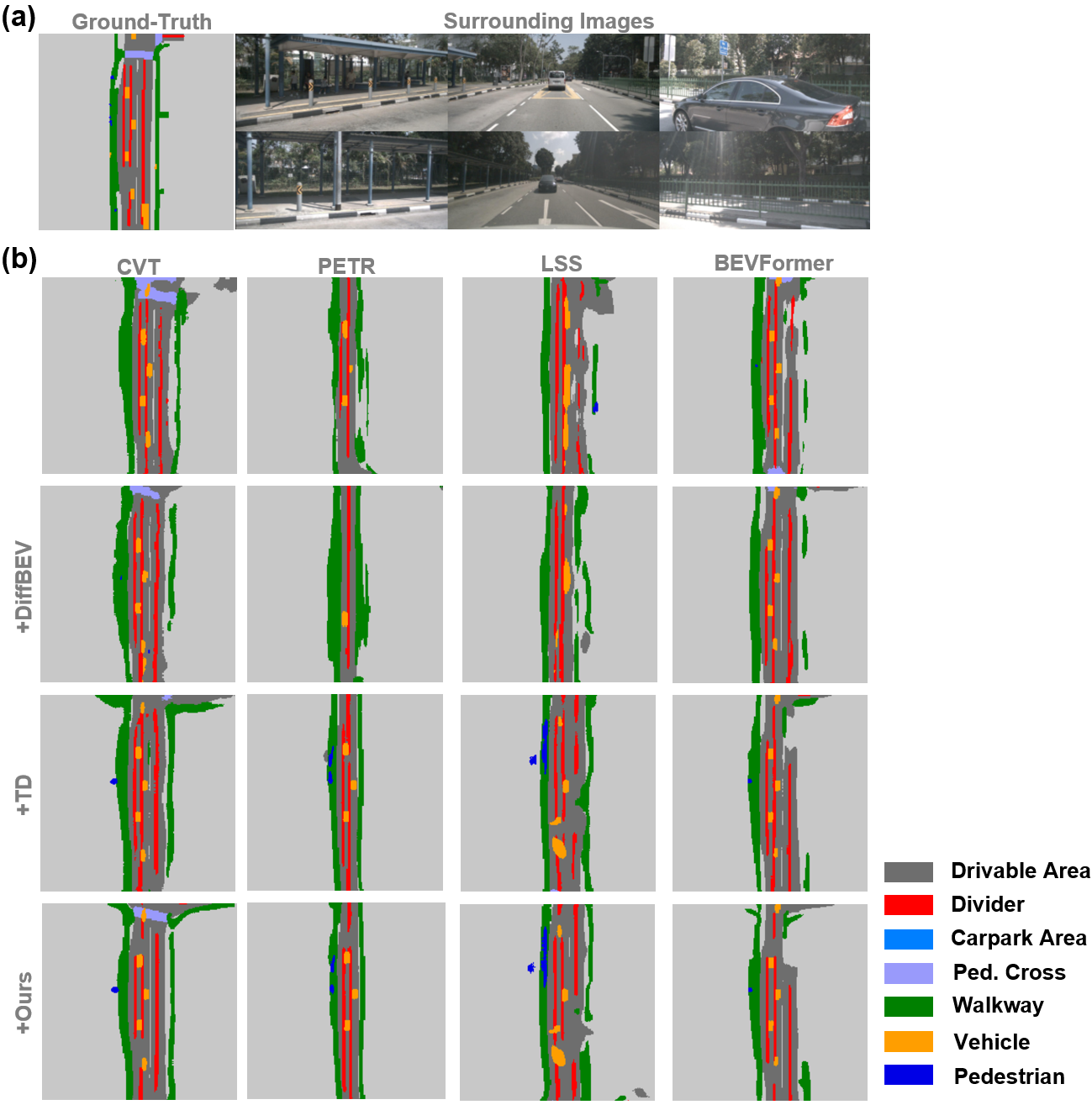}
\caption{Prediction examples.}
\label{sup_fig2}
\end{figure*}

\begin{figure*}[t]
\centering
\includegraphics[width=15.0cm]{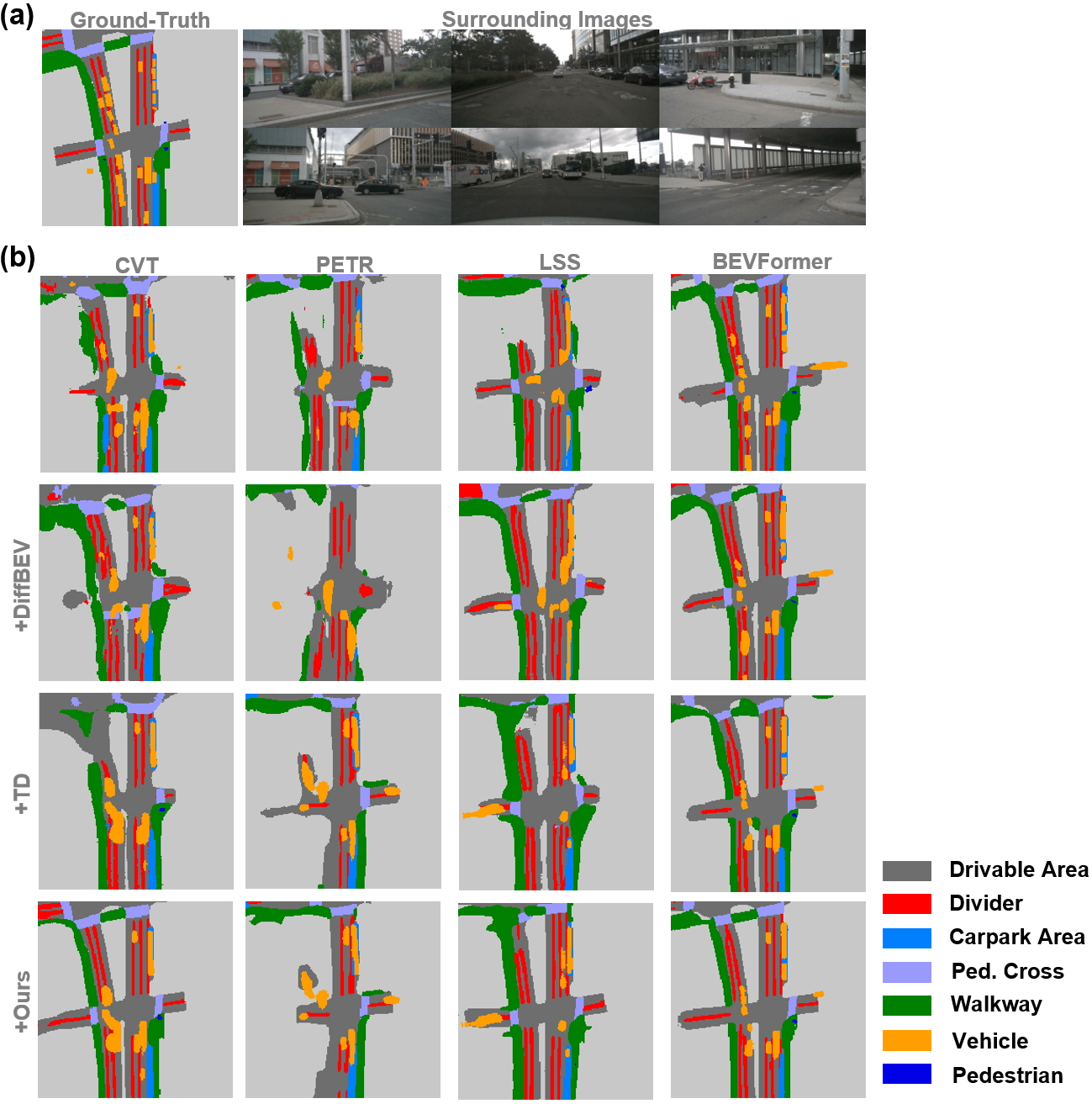}
\caption{Prediction examples.}
\label{sup_fig3}
\end{figure*}

\begin{figure*}[t]
\centering
\includegraphics[width=15.0cm]{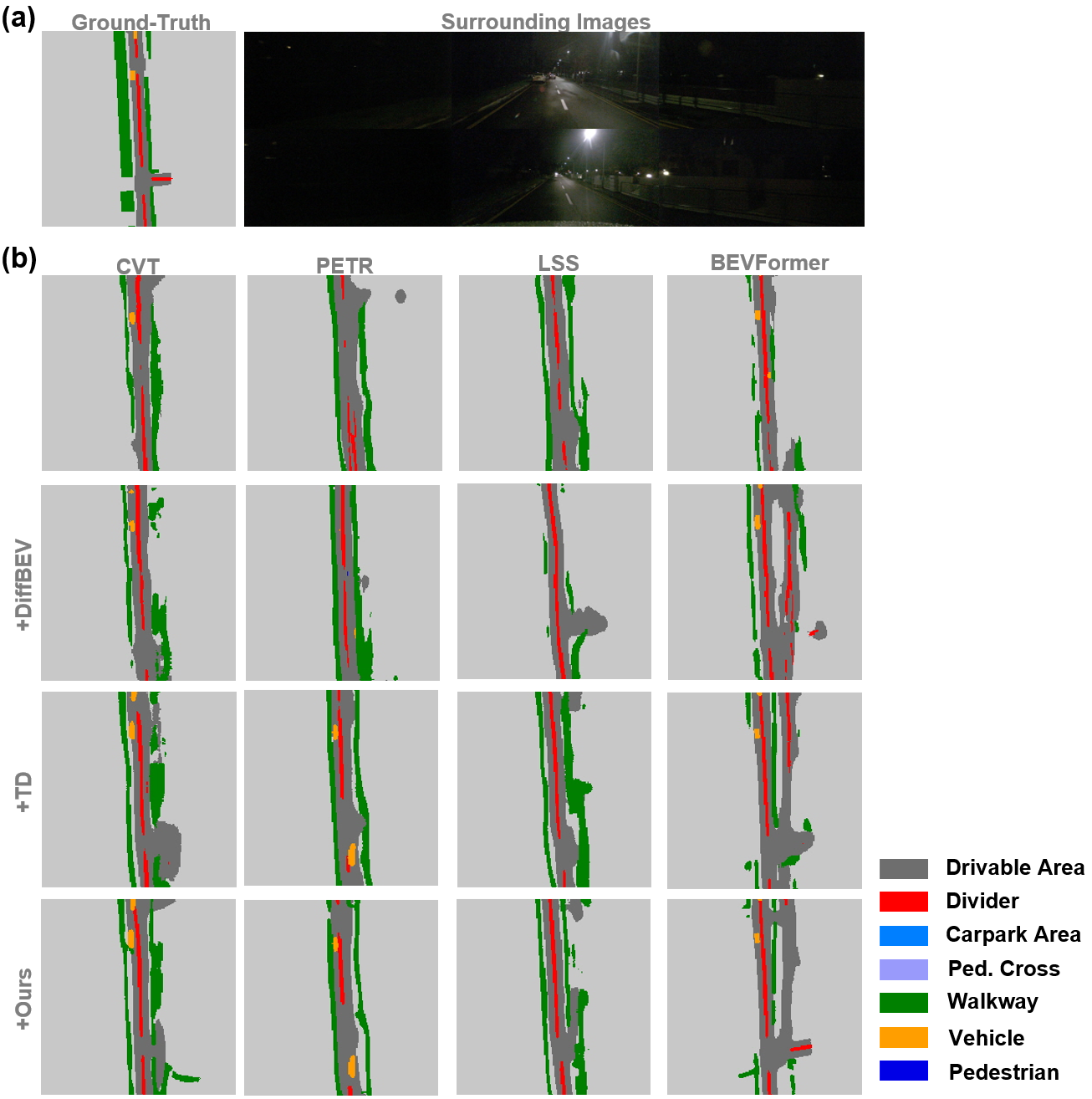}
\caption{Prediction examples.}
\label{sup_fig4}
\end{figure*}

\section{Implementation Details}
\subsection{BEV Map Autoencoder}
Figure \ref{sup_fig1} shows the general architecture of BEV map autoencoder. We trained two autoencoders, one producing $\mathbf{B}^{ae} \in \mathbb{R}^{25 \times 25 \times 256}$ for CVT \cite{Zhou_cvpr22} and PETR \cite{Liu_iccv23} while the other producing $\mathbf{B}^{ae} \in \mathbb{R}^{50 \times 50 \times 256}$ for LSS \cite{Philion_eccv20} and BEVFormer \cite{Li_eccv22}, with the initial learning rate of 5$e^{-4}$. During the training, as proposed in \cite{Zhao_cvpr24}, randomly generated noise of the same size as $\mathbf{B}^{ae}$ is added to $\mathbf{B}^{ae}$ to make the autoencoder robust. The figure shows an example of decoding the map corresponding to $\textit{drivable area}$. If the number of semantic classes is greater than one, additional head and decoder networks dedicated to a specific class is added. For the supervision of the autoencoder, we use binary cross entropy loss as follows:
\begin{equation}
\mathcal{L} = \sum_{c \in \mathcal{C}} \lambda_{c} \cdot BCE(\mathbf{O}_{c},\hat{\mathbf{O}}_{c}),
\label{sup_eqn1}
\end{equation}
where $\mathbf{O}_{c}$ and $\hat{\mathbf{O}}_{c}$ respectively denote the ground-truth BEV map for the class $c$ and its prediction. $\lambda_{c}$ is a pre-defined constant for the class, controlling the influence of $BCE(\mathbf{O}_{c},\hat{\mathbf{O}}_{c})$ on the overall loss. We set $\lambda_{c}$ as shown in Table \ref{sup_tab1} to reflect the fact that the size of the areas occupied by a class in BEV maps differs across the classes. Note that we used $\lambda_{c}$ values in Table \ref{sup_tab1} when training DiffBEV and the four baseline models for a fair comparison.

\subsection{TD Paradigm}
Once the autoencoder is trained at the first stage of the TD paradigm as described above, the parameters of the encoder and head networks are fixed for the subsequent training stages. At the second stage, a VT encoder is trained to produce $\mathbf{B}$ as close as possible to $\mathbf{B}^{ae}$ via $l2$-loss. Finally, the decoders of the autoencoder are fine-tuned at the third stage through Eqn. \ref{sup_eqn1} with $\lambda_{c}=1$ for all classes. When training the noise estimation networks, we train them with the decoders simultaneously through the loss function proposed in the main paper.

\subsection{DiffBEV}
Because the original implementation code of DiffBEV \cite{Zou_aaai24} does not provide the conditional DDPM part for the semantic segmentation task, we implemented it based on the description in the paper. First of all, for a fair comparison, we use the noise estimation network architecture similar to that proposed in our paper. Next, we strictly follow the standard DDPM model to add noise to and predict it from $\mathbf{B}$, while the difference is that condition-modulated denoising is employed as the original paper suggested. We also exclude the depth estimation part of DiffBEV for a fair comparison. With the introduction of the depth estimation part, one can expect better segmentation results.Finally, we incorporated DDIM \cite{Song_iclr21} for fast sampling, and found that the best performance was achieved with four sampling steps.

\subsection{Four Baseline Models}
As we mentioned in the paper, we apply our method to the four representative baseline models: CVT \cite{Zhou_cvpr22}, PETR \cite{Liu_iccv23}, LSS \cite{Philion_eccv20}, and BEVFormer \cite{Li_eccv22}. The implementation codes for the four models are downloaded from the respective github repositories and used for the experiments in this paper. We didn't change the original configurations except the followings: 1) Following the standard practice, the input images are scaled and cropped to 224 × 448 pixels before being fed to an image backbone. 2) We let LSS and BEVFormer produce $\mathbf{B}$ of spatial resolution $50 \times 50$ to make them trainable on our GPUs. 3) We replace the decoders of the four baselines with the adapted version of CVT's decoder to facilitate multi-class prediction in the way that each class is given a dedicated head network and the ConvNet-based original CVT decoder.

\begin{table}[t]
\begin{center}
\scalebox{0.7}{
\begin{tabular}{|c|c|}
\hline
Class                       & $\lambda_{c}$\\
\hline
\textit{Drivable Area}      & 0.03  \\
\textit{Divider}            & 0.25 \\
\textit{Ped. Cross}      & 0.25 \\
\textit{CarPark Area}       & 0.25 \\
\textit{Walkway}            & 0.25 \\
\textit{Vehicle}            & 0.5 \\
\textit{Pedestrian}         & 1.0  \\
\hline

\end{tabular}
}
\end{center}
\caption{$\lambda_{c}$ for each class.}
\label{sup_tab1}
\end{table}


\end{document}